\documentclass[letterpaper]{article}
\usepackage[T1]{fontenc}
\usepackage{spconf,amsmath,amssymb,graphicx,booktabs,float,placeins,cite,xurl,hyperref}
\hypersetup{hidelinks}
\graphicspath{{figures/}}
\newcommand{\gtranet}{G\textsuperscript{2}RA-Net}
\newcommand{\R}{\mathbb{R}}

\title{G\textsuperscript{2}RA-NET: GRAPH-BASED CROSS-SLICE RELATION MODELING WITH ATTENTION GATING FOR MEDICAL IMAGE SEGMENTATION}
\name{Shengye Wang\textsuperscript{1}, Zonglin Wu\textsuperscript{1}\sthanks{Corresponding author: Zonglin Wu. This work has been submitted to the IEEE for possible publication. Copyright may be transferred without notice, after which this version may no longer be accessible.}, Liang Fan\textsuperscript{2}, Yule Xue\textsuperscript{1}, Haozhe Zhao\textsuperscript{3}}
\address{\textsuperscript{1}College of Computer and Information Science, Southwest University, Chongqing, China\\
\textsuperscript{2}ai.io, London, UK\qquad
\textsuperscript{3}Chengyi College, Jimei University, Xiamen, China}

\begin{document}
\ninept
\maketitle

\begin{abstract}
Medical image segmentation supports quantitative clinical analysis and computer-aided diagnosis. Recent methods for medical image segmentation have improved both local feature representation and volumetric context modeling. However, existing methods still struggle to efficiently model cross-slice relations in anisotropic volumetric images, limiting segmentation consistency and accuracy. This paper proposes \gtranet{}, a medical image segmentation framework that combines graph-based cross-slice relation modeling with attention gating. Graph-Based Slice Relationship Modeling (GSRM) captures anatomical dependencies across consecutive slices by representing each slice as a graph node and propagating semantic context through graph message passing. The Cross-Slice Attention Gate (CSAG) then selects relevant neighboring context and emphasizes target anatomical regions through attention-guided feature modulation. Experiments on brain MRI and abdominal CT datasets demonstrate that \gtranet{} outperforms representative methods in segmentation accuracy and boundary quality. Ablation studies further validate the proposed design.
\end{abstract}


\section{Introduction}
\label{sec:introduction}

Medical image segmentation supports quantitative clinical analysis and computer-aided diagnosis by providing anatomical and pathological information for measurement, disease assessment, and treatment planning~\cite{ref01,ref24}. Accurate and efficient delineation of organs and lesions is essential for reliable analysis and timely clinical decisions. However, manual delineation is time-consuming and labor-intensive~\cite{ref26}. Complex anatomical structures, ambiguous boundaries, and the need to maintain consistency across neighboring slices further complicate accurate segmentation. Clinical experience and subjective judgment can also affect results~\cite{ref02}. Advances in deep learning offer a new way to overcome these bottlenecks~\cite{ref03,ref04}.

Recent deep-learning methods have advanced medical image segmentation in areas including local feature representation and volumetric context modeling~\cite{ref03,ref04,ref05,ref06,ref07}. Local feature representation preserves fine anatomical details and accurate boundaries, while volumetric context modeling maintains anatomical continuity and consistent predictions across slices. Encoder--decoder-based segmentation methods have been effective in both aspects. For local feature representation, the encoder extracts multi-level representations at decreasing spatial resolutions and increasing receptive fields, while the decoder restores spatial resolution for dense pixel-level prediction. Skip connections fuse encoder and decoder features at corresponding scales, recovering high-resolution information lost during downsampling~\cite{ref03}. For volumetric medical image data, these methods differ in how they use spatial information: some process individual slices independently, whereas others jointly model multiple consecutive slices. The former preserve fine in-plane details at relatively low computational cost, while the latter capture richer volumetric context~\cite{ref04,ref26}. These advantages have made encoder--decoder frameworks widely adopted in medical image segmentation~\cite{ref25}.

Efficient use of cross-slice context remains challenging. Slice-wise 2D methods are efficient but omit inter-slice dependencies, whereas 3D models capture volumetric context at greater computational and memory cost~\cite{ref26} and can struggle with in-plane/through-plane anisotropy~\cite{ref05}. Cross-slice attention provides a 2.5D alternative~\cite{ref05,ref06,ref07}. However, slice features vary in their relevance to the segmentation target, making selective weighting important for suppressing false-positive responses~\cite{ref05}. These challenges motivate explicit slice-relation modeling with selective context use within a 2D backbone.

To address these challenges, we propose \gtranet{}, which combines cross-slice relation modeling and attention gating within a 2D U-Net backbone~\cite{ref03}. \gtranet{} introduces two complementary modules: Graph-Based Slice Relationship Modeling (GSRM) and the Cross-Slice Attention Gate (CSAG). At the bottleneck, GSRM represents consecutive slices as graph nodes and uses an adjacency structure to encode their anatomical relationships. Lightweight graph message passing then propagates slice-level semantic context, which modulates the bottleneck features to incorporate cross-slice anatomical information. CSAG selectively integrates relevant neighboring context and emphasizes target anatomical regions by combining cross-slice self-attention, a slice-adjacency prior, and spatial attention at the bottleneck and each decoder stage. Experiments on the L2R-OASIS brain MRI and L2R-Abdomen CT datasets show that \gtranet{} outperforms representative methods in segmentation accuracy and boundary quality. Ablation studies further validate the proposed design.

\section{Methods}
\label{sec:methods}

\subsection{Overview}
\label{subsec:overview}

\begin{figure}[!t]
  \centering
  \includegraphics[width=\columnwidth,clip]{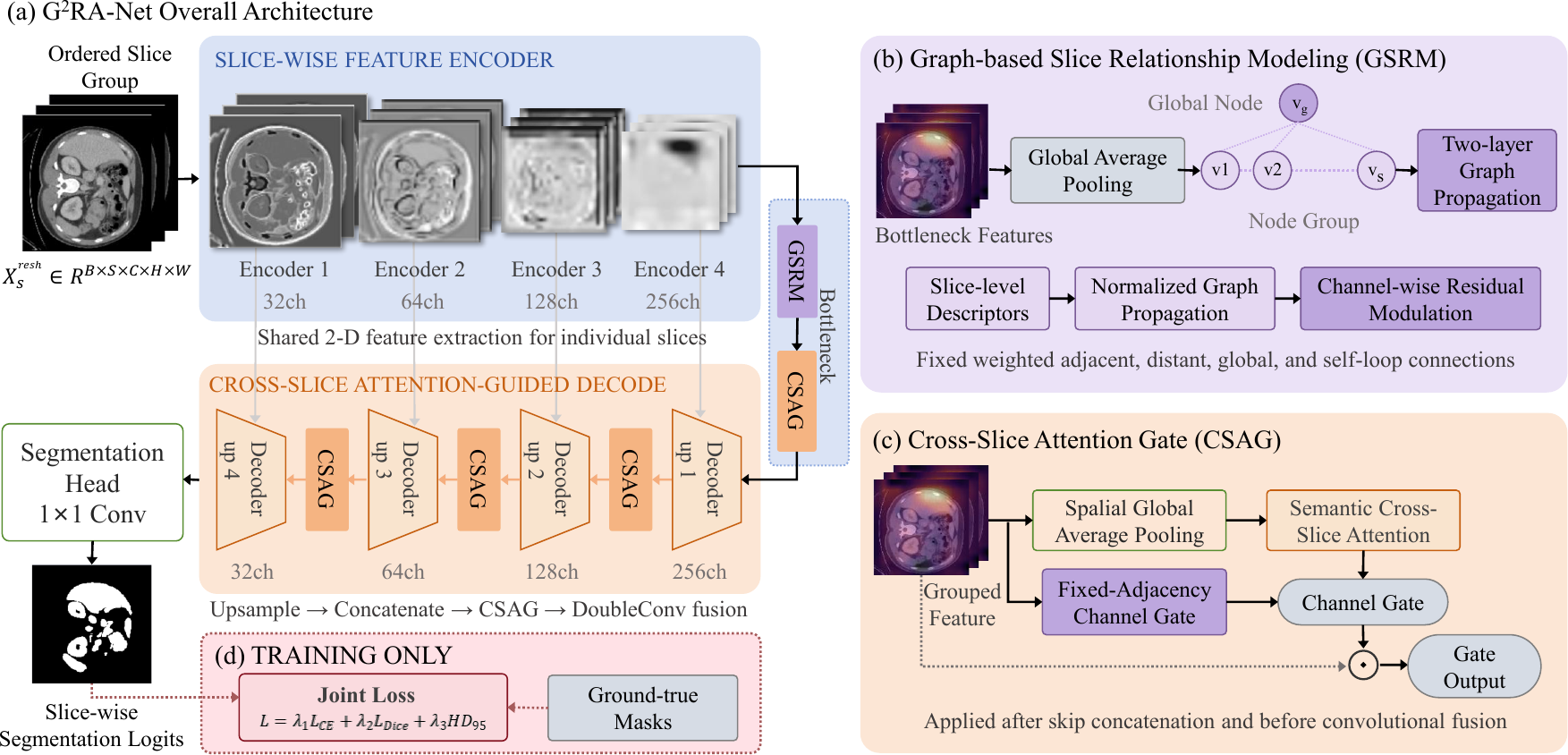}
  \caption{Overall architecture of \gtranet. GSRM models cross-slice relations at the bottleneck, while CSAG gates bottleneck and decoder features during reconstruction.}
  \label{fig:overall}
\end{figure}

Fig.~\ref{fig:overall} shows the overall architecture of \gtranet{}, a 2D U-Net encoder-decoder~\cite{ref03} that takes as input an ordered group of consecutive slices from the same volume. The encoder independently processes each slice through successive 2D convolution and downsampling stages to extract multi-scale spatial features. At the bottleneck, these features are reshaped according to slice order and fed into Graph-Based Slice Relationship Modeling (GSRM). GSRM treats each slice as a graph node and obtains its node feature through global average pooling. It then performs lightweight message propagation between adjacent nodes, and the updated node features modulate the bottleneck maps. The Cross-Slice Attention Gate (CSAG) then selects relevant cross-slice information and suppresses inconsistent bottleneck responses. At each decoder stage, CSAG modulates the concatenated upsampled and corresponding encoder skip features before convolutional fusion. Finally, a $1\times1$ convolution produces slice-wise segmentation logits.

\subsection{Graph-Based Slice Relationship Modeling (GSRM)}
\label{subsec:gsrm}

\begin{figure}[!t]
  \centering
  \includegraphics[width=\columnwidth,clip]{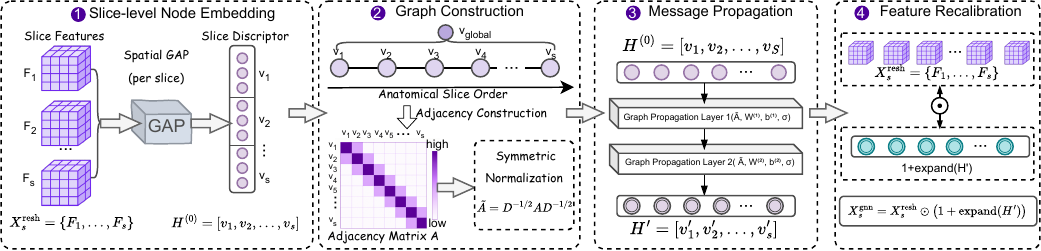}
  \caption{Graph-Based Slice Relationship Modeling, including slice-level node embedding, graph construction, message propagation, and feature recalibration.}
  \label{fig:gsrm}
\end{figure}

GSRM enables each slice to acquire semantic context from adjacent slices at the bottleneck without changing the 2D convolutional backbone. GSRM constructs slice nodes, propagates information over the slice graph, and modulates the bottleneck features through residual channel scaling, as illustrated in Fig.~\ref{fig:gsrm}. Given the reshaped bottleneck feature $X_{5}^{\mathrm{resh}}\in\R^{B\times S\times C\times H\times W}$, where $B$ and $S$ denote the number of volumes and consecutive slices, respectively, each slice is treated as a graph node, and global average pooling produces the node feature tensor $H\in\R^{B\times S\times C}$, with each node feature computed as:
\begin{equation}
h_{b,s}=\frac{1}{HW}\sum_{i=1}^{H}\sum_{j=1}^{W}X_{5,b,s,:,i,j}.
\label{eq:node-feature}
\end{equation}
After constructing the node features, an adjacency matrix $A$ describes the connections between slice nodes and is symmetrically normalized as
\begin{equation}
\widetilde{A}=D^{-1/2}AD^{-1/2},
\label{eq:adjacency}
\end{equation}
where $D$ is the degree matrix. Two graph-propagation layers with learnable linear transformations and nonlinear activations perform message propagation over the normalized adjacency matrix, yielding the updated node features $H'\in\R^{B\times S\times C}$. Rather than replacing the spatial feature maps, GSRM expands $H'$ back to the spatial dimensions and applies residual channel modulation:
\begin{equation}
X_{5}^{\mathrm{gnn}}=X_{5}^{\mathrm{resh}}\odot\left(1+\operatorname{expand}(H')\right).
\label{eq:gsrm-modulation}
\end{equation}
Here, $\operatorname{expand}(\cdot)$ extends the node features from $B\times S\times C$ to $B\times S\times C\times H\times W$. Each slice thus obtains a graph-propagated channel scaling factor that incorporates cross-slice semantic context into the bottleneck features while preserving their spatial layout. Because the slice graph contains only $S$ nodes, graph propagation incurs little overhead while explicitly encoding the prior that adjacent slices are more closely related, thereby enabling more stable identification of axially continuous structures.

\subsection{Cross-Slice Attention Gate (CSAG)}
\label{subsec:csag}

\begin{figure}[!t]
  \centering
  \includegraphics[width=\columnwidth,clip]{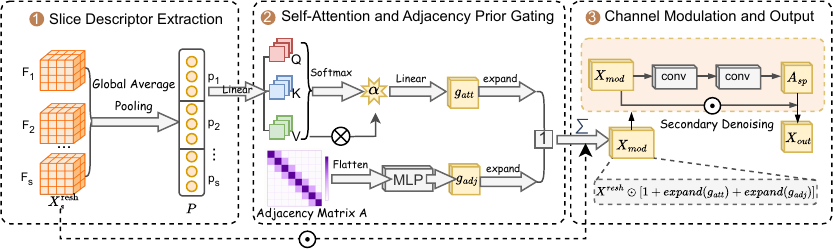}
  \caption{Cross-Slice Attention Gate, comprising slice descriptor extraction, self-attention and adjacency-prior gating, channel modulation, and spatial refinement.}
  \label{fig:csag}
\end{figure}

Graph propagation at the bottleneck alone may not fully constrain multiscale detail recovery. \gtranet{} therefore applies the Cross-Slice Attention Gate (CSAG) at the bottleneck and every decoder stage to select relevant cross-slice information and suppress irrelevant feature responses during multiscale reconstruction. CSAG performs cross-slice channel modulation followed by within-slice spatial refinement, as illustrated in Fig.~\ref{fig:csag}. Given $X\in\R^{(B\cdot S)\times C\times H\times W}$, CSAG restores the slice dimension to obtain $X^{\mathrm{resh}}\in\R^{B\times S\times C\times H\times W}$. Global average pooling produces slice descriptors $P\in\R^{B\times S\times C}$. Linear projections generate $Q$, $K$, and $V$, and slice-wise self-attention weights~\cite{ref08} are computed as
\begin{equation}
\alpha=\operatorname{softmax}\!\left(\frac{QK^{T}}{\sqrt{d}}\right)\in\R^{B\times S\times S}.
\label{eq:self-attention}
\end{equation}
Here, $d$ is the attention dimension. $Q$ and $K$ estimate pairwise relevance between slices, whereas $V$ carries the information to be aggregated across slices. The attention weights aggregate $V$ to obtain $Z=\alpha V\in\R^{B\times S\times d}$. After projection back to the channel dimension, $Z$ yields $g_{\mathrm{att}}\in\R^{B\times S\times C}$, a content-dependent slice-wise channel gating vector that strengthens channels supported by semantically related slices and suppresses inconsistent responses. In parallel, flattening $A\in\R^{B\times S\times S}$ and mapping it through a multilayer perceptron (MLP) produces the adjacency gating vector $g_{\mathrm{adj}}\in\R^{B\times S\times C}$. This vector explicitly encodes the structural prior that adjacent slices are more closely related and controls how adjacent slices influence semantic channels. The two gating vectors jointly perform residual channel modulation:
\begin{equation}
\begin{aligned}
X^{\mathrm{mod}}=X^{\mathrm{resh}}\odot\bigl(&1+\operatorname{expand}(g_{\mathrm{att}})\\
&+\operatorname{expand}(g_{\mathrm{adj}})\bigr).
\end{aligned}
\label{eq:csag-channel}
\end{equation}
On this basis, the module reshapes $X^{\mathrm{mod}}$ by merging $B$ and $S$ into the batch dimension and then uses two convolutional layers to generate a spatial attention map $A_{\mathrm{sp}}\in\R^{(B\cdot S)\times1\times H\times W}$. The spatial attention map weights each spatial location to emphasize task-relevant responses and suppress irrelevant within-slice activations:
\begin{equation}
X^{\mathrm{out}}=X^{\mathrm{mod}}\odot A_{\mathrm{sp}}.
\label{eq:csag-output}
\end{equation}
At the bottleneck, CSAG further aligns cross-slice semantic responses. At each decoder stage, CSAG acts on the concatenated upsampled and skip features before convolutional fusion. The gate therefore controls both the details carried by skip connections and their combination with upsampled features throughout multiscale reconstruction.

\subsection{Loss Function and Evaluation Metrics}
\label{subsec:loss-metrics}

Training jointly optimizes pixel-wise classification, region overlap, and boundary accuracy using cross-entropy loss $L_{\mathrm{CE}}$, Dice loss $L_{\mathrm{Dice}}$~\cite{ref09}, and differentiable Hausdorff distance-transform loss $L_{\mathrm{HDT}}$~\cite{ref10}:
\begin{equation}
L=\lambda_{1}L_{\mathrm{CE}}+\lambda_{2}L_{\mathrm{Dice}}+\lambda_{3}L_{\mathrm{HDT}},
\label{eq:loss}
\end{equation}
where $\lambda_{1}$, $\lambda_{2}$, and $\lambda_{3}$ are weighting coefficients for $L_{\mathrm{CE}}$, $L_{\mathrm{Dice}}$, and $L_{\mathrm{HDT}}$, respectively.

We use Macro Dice and Macro IoU to assess region overlap and Macro HD95 and Macro ASD to assess boundary error, combining complementary metric families as our primary measures~\cite{ref11,ref12}. Macro Precision and Macro Recall assess foreground classification, while Micro Dice and frequency-weighted Dice (FW Dice) summarize overall pixel-level agreement. Pooled pixel-level scores can mask errors in small structures~\cite{ref27}. We therefore calculate macro scores by averaging over ground-truth-present foreground classes within each case and then across cases, giving equal weight to eligible classes within each case and to individual cases. Higher overlap scores and lower distance values indicate better segmentation.

\raggedbottom
\setlength{\floatsep}{3pt plus 1pt minus 1pt}
\setlength{\textfloatsep}{4pt plus 1pt minus 1pt}
\setlength{\dbltextfloatsep}{4pt plus 1pt minus 1pt}
\makeatletter
\long\def\@makecaption#1#2{%
 \vskip 5pt
 \setbox\@tempboxa\hbox{#1. #2}%
 \ifdim \wd\@tempboxa >\hsize #1. #2\par \else \hbox
 to\hsize{\hfil\box\@tempboxa\hfil}\fi}
\makeatother
\section{Experiments}
\label{sec:experiments}

\subsection{Dataset}
\label{subsec:dataset}

We evaluate \gtranet{} on L2R-OASIS 1.1~\cite{ref13,ref14} (414 brain MRI volumes, 35 foreground structures) and L2R-Abdomen CT-CT 1.1~\cite{ref14} (30 abdominal CT volumes, 13 foreground organs). Their training/validation/test splits are 290/62/62 and 21/5/4, respectively.

\subsection{Implementation Details}
\label{subsec:implementation}

Training uses PyTorch on an NVIDIA GeForce RTX 4080 (16 GB), with three consecutive axial slices (stride 1) per input. Images and masks are resized to $224\times224$ using bilinear and nearest-neighbor interpolation, respectively. MRI is min--max normalized per volume; CT intensities are clipped to $[-175,250]$ HU and normalized to $[0,1]$. We train for up to 100 epochs with batch size 2 using AdamW~\cite{ref15} (initial learning rate $10^{-4}$, weight decay $10^{-4}$). Testing uses the checkpoint with the highest validation Macro Dice. Unless specified otherwise, experiments share splits, preprocessing, and evaluation.

\subsection{Results on the Two Datasets}
\label{subsec:results}

\begin{figure}[!t]
  \centering
  \includegraphics[width=\columnwidth]{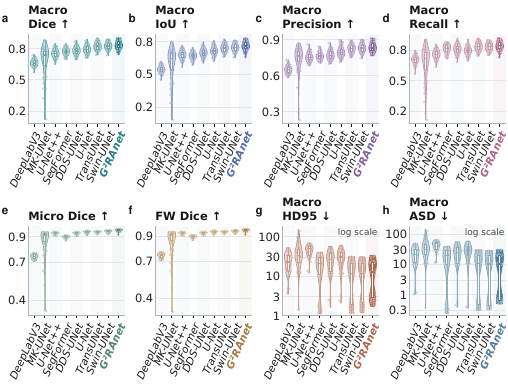}
  \caption{L2R-OASIS comparison (62 volumes): violins show distributions, dots show cases, and white boxes show interquartile ranges with median lines.}
  \label{fig:oasis-external}
\end{figure}

\begin{figure}[!t]
  \centering
  \includegraphics[width=\columnwidth]{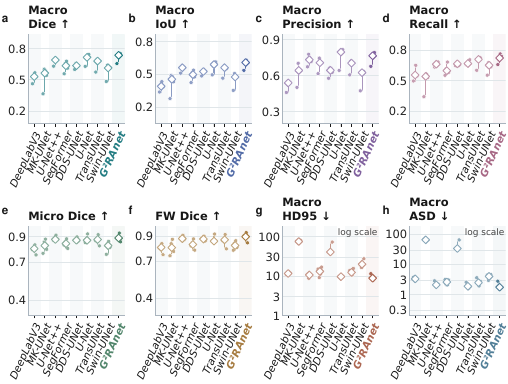}
  \caption{L2R-Abdomen CT comparison (four volumes): dots show cases, diamonds show medians, and lines show observed ranges.}
  \label{fig:abdomen-external}
\end{figure}

Figs.~\ref{fig:oasis-external} and~\ref{fig:abdomen-external} compare \gtranet{} with eight representative methods~\cite{ref17,ref18,ref19,ref20,ref21,ref22,ref23,ref03} on L2R-OASIS and L2R-Abdomen CT. \gtranet{} achieves the highest mean Macro Dice and Macro IoU and the lowest mean Macro HD95 and Macro ASD on both datasets. The figures show case-level distributions with medians, while the comparisons below use case-averaged scores. Each test volume contributes equally to the reported means. HD95 and ASD are plotted on logarithmic scales.

On L2R-OASIS, \gtranet{} achieves Macro Dice of 0.8328 and Macro IoU of 0.7603, improving over Swin-UNet, the strongest baseline on both metrics, by 0.84 and 1.70 percentage points, respectively. Against the same baseline, Macro HD95 decreases from 13.08 to 12.43 and Macro ASD from 11.83 to 10.25. Micro Dice and FW Dice also exceed those of all compared baselines. These gains cover class-averaged overlap, pixel-level agreement, and boundary accuracy.

On L2R-Abdomen CT, \gtranet{} outperforms U-Net, the strongest external baseline on both overlap and boundary metrics. Macro Dice increases from 0.6999 to 0.7174 and Macro IoU from 0.5709 to 0.5939, corresponding to gains of 1.75 and 2.30 percentage points. Macro HD95 decreases from 9.49 to 9.16 and Macro ASD from 1.96 to 1.93. Macro Precision and Macro Recall also improve over U-Net, indicating that the overlap gains are accompanied by improved foreground classification.

\begin{figure*}[!t]
  \centering
  \includegraphics[width=0.92\textwidth]{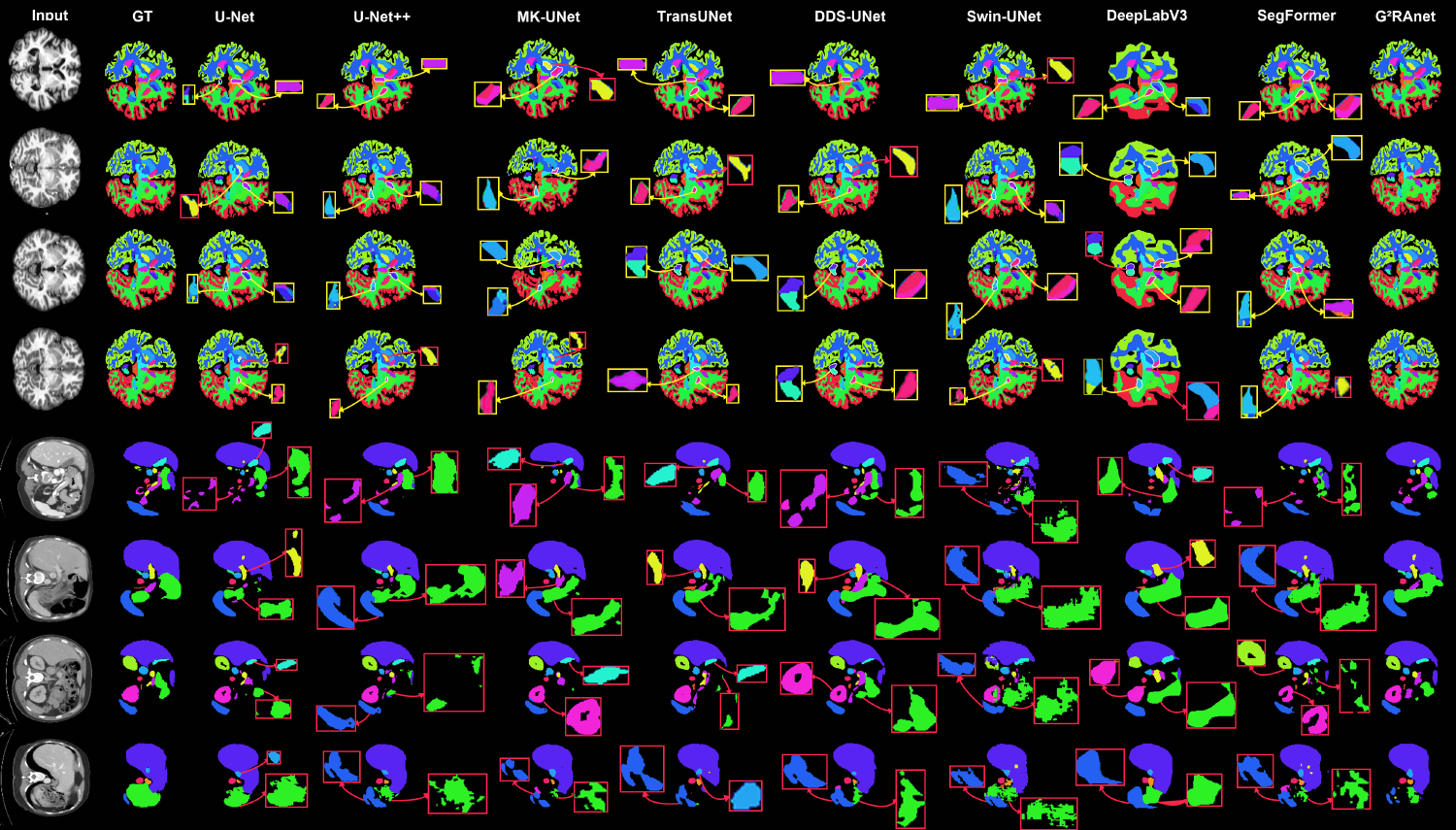}
  \caption{Qualitative comparison of segmentation results on the
  L2R-OASIS and L2R-Abdomen CT datasets.}
  \label{fig:qualitative}
\end{figure*}

Tables~\ref{tab:oasis-macro-ablation}--\ref{tab:abdomen-pixel-boundary-ablation} assess the incremental effects of adding GSRM and then CSAG to U-Net. On OASIS, GSRM yields the larger Macro Dice gain (3.10 percentage points), followed by a further 1.34-point gain from CSAG. Macro HD95 decreases from 30.00 to 13.48 and then to 12.43, while Macro ASD decreases from 28.93 to 12.33 and then to 10.25. Macro IoU increases from 0.7125 to 0.7428 and then to 0.7603, corresponding to gains of 3.03 and 1.75 percentage points. Both overlap measures therefore show a larger incremental gain from GSRM on OASIS.

On Abdomen CT, CSAG yields the larger incremental Macro Dice gain (1.75 versus 0.77 percentage points for GSRM). CSAG also yields the larger Macro IoU gain (2.30 versus 0.36 percentage points), so both overlap measures favor the second addition. Together, the two additions raise Macro Dice from 0.6922 to 0.7174, a total gain of 2.52 percentage points. Each addition also improves foreground classification and pixel-level agreement and reduces boundary error.

\begingroup
\setlength{\intextsep}{2pt plus 1pt minus 1pt}
\begin{table}[!t]
\caption{L2R-OASIS ablation: macro-averaged segmentation metrics.}
\label{tab:oasis-macro-ablation}
\centering
\renewcommand{\arraystretch}{1.00}
\setlength{\tabcolsep}{3pt}
\begin{tabular*}{\columnwidth}{@{\extracolsep{\fill}}lrrrr@{}}
\toprule
method & \multicolumn{1}{c}{\shortstack{Macro\\Dice\_\\mean $\uparrow$}} & \multicolumn{1}{c}{\shortstack{Macro\\IoU\_\\mean $\uparrow$}} & \multicolumn{1}{c}{\shortstack{Macro\\Precision\_\\mean $\uparrow$}} & \multicolumn{1}{c}{\shortstack{Macro\\Recall\_\\mean $\uparrow$}} \\
\midrule
U-Net & 0.7884 & 0.7125 & 0.7994 & 0.7922 \\
U-Net + GSRM & 0.8194 & 0.7428 & 0.8227 & 0.8319 \\
\gtranet & \textbf{0.8328} & \textbf{0.7603} & \textbf{0.8343} & \textbf{0.8449} \\
\bottomrule
\end{tabular*}
\end{table}

\begin{table}[!t]
\caption{L2R-OASIS ablation: pixel-level agreement and boundary error.}
\label{tab:oasis-pixel-boundary-ablation}
\centering
\renewcommand{\arraystretch}{1.00}
\setlength{\tabcolsep}{3pt}
\begin{tabular*}{\columnwidth}{@{\extracolsep{\fill}}lrrrr@{}}
\toprule
method & \multicolumn{1}{c}{\shortstack{Micro\\Dice $\uparrow$}} & \multicolumn{1}{c}{\shortstack{FW\\Dice $\uparrow$}} & \multicolumn{1}{c}{\shortstack{Macro\\HD95 $\downarrow$}} & \multicolumn{1}{c}{\shortstack{Macro\\ASD $\downarrow$}} \\
\midrule
U-Net & 0.9346 & 0.9343 & 30.00 & 28.93 \\
U-Net + GSRM & 0.9445 & 0.9444 & 13.48 & 12.33 \\
\gtranet & \textbf{0.9473} & \textbf{0.9473} & \textbf{12.43} & \textbf{10.25} \\
\bottomrule
\end{tabular*}
\end{table}

\begin{table}[!t]
\caption{L2R-Abdomen CT ablation: macro-averaged segmentation metrics.}
\label{tab:abdomen-macro-ablation}
\centering
\renewcommand{\arraystretch}{1.00}
\setlength{\tabcolsep}{3pt}
\begin{tabular*}{\columnwidth}{@{\extracolsep{\fill}}lrrrr@{}}
\toprule
method & \multicolumn{1}{c}{\shortstack{Macro\\Dice\_\\mean $\uparrow$}} & \multicolumn{1}{c}{\shortstack{Macro\\IoU\_\\mean $\uparrow$}} & \multicolumn{1}{c}{\shortstack{Macro\\Precision\_\\mean $\uparrow$}} & \multicolumn{1}{c}{\shortstack{Macro\\Recall\_\\mean $\uparrow$}} \\
\midrule
U-Net & 0.6922 & 0.5673 & 0.7322 & 0.6746 \\
U-Net + GSRM & 0.6999 & 0.5709 & 0.7496 & 0.6900 \\
\gtranet & \textbf{0.7174} & \textbf{0.5939} & \textbf{0.7629} & \textbf{0.7164} \\
\bottomrule
\end{tabular*}
\end{table}

\begin{table}[!t]
\caption{L2R-Abdomen CT ablation: pixel-level agreement and boundary error.}
\label{tab:abdomen-pixel-boundary-ablation}
\centering
\renewcommand{\arraystretch}{1.00}
\setlength{\tabcolsep}{3pt}
\begin{tabular*}{\columnwidth}{@{\extracolsep{\fill}}lrrrr@{}}
\toprule
method & \multicolumn{1}{c}{\shortstack{Micro\\Dice $\uparrow$}} & \multicolumn{1}{c}{\shortstack{FW\\Dice $\uparrow$}} & \multicolumn{1}{c}{\shortstack{Macro\\HD95 $\downarrow$}} & \multicolumn{1}{c}{\shortstack{Macro\\ASD $\downarrow$}} \\
\midrule
U-Net & 0.8770 & 0.8721 & 9.49 & 1.96 \\
U-Net + GSRM & 0.8839 & 0.8833 & 9.41 & 1.95 \\
\gtranet & \textbf{0.8890} & \textbf{0.8909} & \textbf{9.16} & \textbf{1.93} \\
\bottomrule
\end{tabular*}
\end{table}

\endgroup

Fig.~\ref{fig:qualitative} compares the predicted segmentations with the ground truth in selected MRI and CT cases. \gtranet{} preserves more complete fine structures in brain MRI and follows abdominal organ contours more closely, with fewer fragmented or missing regions. These observations complement the quantitative overlap and boundary results.

\FloatBarrier
\vspace{-5pt}
\section{Conclusion}
\label{sec:conclusion}
\vspace{-1pt}

Across L2R-OASIS and L2R-Abdomen CT, \gtranet{} achieves the highest mean Macro Dice and Macro IoU and the lowest mean Macro HD95 and Macro ASD among the evaluated methods. Ablation results show complementary gains from GSRM and CSAG across the two datasets: GSRM yields the larger incremental Macro Dice gain on L2R-OASIS, whereas the subsequent addition of CSAG yields the larger gain on L2R-Abdomen CT. Qualitative comparisons further show more complete structures, fewer fragmented or missing regions, and clearer anatomical boundaries for the complete model in the selected cases. Together, these findings support the effectiveness of the two-module design for cross-slice medical image segmentation.


\begin{thebibliography}{26}
\bibitem{ref01}
E. Gibbons, M. Hoffmann, J. Westhuyzen, A. Hodgson, B. Chick, and A. Last, ``Clinical evaluation of deep learning and atlas-based auto-segmentation for critical organs at risk in radiation therapy,'' \emph{J. Med. Radiat. Sci.}, vol. 70, suppl. 2, pp. 15--25, 2023, doi: 10.1002/jmrs.618.

\bibitem{ref24}
G. Litjens et al., ``A survey on deep learning in medical image analysis,'' \emph{Med. Image Anal.}, vol. 42, pp. 60--88, 2017, doi: 10.1016/j.media.2017.07.005.

\bibitem{ref26}
X. Li, H. Chen, X. Qi, Q. Dou, C.-W. Fu, and P.-A. Heng, ``H-DenseUNet: Hybrid densely connected UNet for liver and tumor segmentation from CT volumes,'' \emph{IEEE Trans. Med. Imaging}, vol. 37, no. 12, pp. 2663--2674, 2018, doi: 10.1109/TMI.2018.2845918.

\bibitem{ref02}
G. Chlebus et al., ``Reducing inter-observer variability and interaction time of MR liver volumetry by combining automatic CNN-based liver segmentation and manual corrections,'' \emph{PLoS ONE}, vol. 14, no. 5, Art. no. e0217228, 2019, doi: 10.1371/journal.pone.0217228.

\bibitem{ref03}
O. Ronneberger, P. Fischer, and T. Brox, ``U-Net: Convolutional networks for biomedical image segmentation,'' in \emph{Proc. Med. Image Comput. Comput.-Assist. Intervent. (MICCAI)}, 2015, pp. 234--241, doi: 10.1007/978-3-319-24574-4\_28.

\bibitem{ref04}
\"{O}. \c{C}i\c{c}ek, A. Abdulkadir, S. S. Lienkamp, T. Brox, and O. Ronneberger, ``3D U-Net: Learning dense volumetric segmentation from sparse annotation,'' in \emph{Proc. Med. Image Comput. Comput.-Assist. Intervent. (MICCAI)}, 2016, pp. 424--432, doi: 10.1007/978-3-319-46723-8\_49.

\bibitem{ref05}
A. L. Y. Hung et al., ``CSAM: A 2.5D cross-slice attention module for anisotropic volumetric medical image segmentation,'' in \emph{Proc. IEEE/CVF Winter Conf. Appl. Comput. Vis. (WACV)}, 2024, pp. 5911--5920, doi: 10.1109/WACV57701.2024.00582.

\bibitem{ref06}
A. Kumar et al., ``A flexible 2.5D medical image segmentation approach with in-slice and cross-slice attention,'' \emph{Comput. Biol. Med.}, vol. 182, Art. no. 109173, 2024, doi: 10.1016/j.compbiomed.2024.109173.

\bibitem{ref07}
A. L. Y. Hung, H. Zheng, Q. Miao, S. S. Raman, D. Terzopoulos, and K. Sung, ``CAT-Net: A cross-slice attention transformer model for prostate zonal segmentation in MRI,'' \emph{IEEE Trans. Med. Imaging}, vol. 42, no. 1, pp. 291--303, 2023, doi: 10.1109/TMI.2022.3211764.

\bibitem{ref25}
F. Isensee, P. F. Jaeger, S. A. A. Kohl, J. Petersen, and K. H. Maier-Hein, ``nnU-Net: A self-configuring method for deep learning-based biomedical image segmentation,'' \emph{Nat. Methods}, vol. 18, no. 2, pp. 203--211, 2021, doi: 10.1038/s41592-020-01008-z.

\bibitem{ref08}
A. Vaswani et al., ``Attention is all you need,'' in \emph{Adv. Neural Inf. Process. Syst.}, 2017, vol. 30, pp. 5998--6008. [Online]. Available: \url{https://proceedings.neurips.cc/paper/2017/hash/3f5ee243547dee91fbd053c1c4a845aa-Abstract.html}

\bibitem{ref09}
F. Milletari, N. Navab, and S.-A. Ahmadi, ``V-Net: Fully convolutional neural networks for volumetric medical image segmentation,'' in \emph{Proc. 4th Int. Conf. 3D Vis. (3DV)}, 2016, pp. 565--571, doi: 10.1109/3DV.2016.79.

\bibitem{ref10}
D. Karimi and S. E. Salcudean, ``Reducing the Hausdorff distance in medical image segmentation with convolutional neural networks,'' \emph{IEEE Trans. Med. Imaging}, vol. 39, no. 2, pp. 499--513, 2020, doi: 10.1109/TMI.2019.2930068.

\bibitem{ref11}
A. A. Taha and A. Hanbury, ``Metrics for evaluating 3D medical image segmentation: Analysis, selection, and tool,'' \emph{BMC Med. Imaging}, vol. 15, Art. no. 29, 2015, doi: 10.1186/s12880-015-0068-x.

\bibitem{ref12}
L. Maier-Hein et al., ``Metrics reloaded: Recommendations for image analysis validation,'' \emph{Nat. Methods}, vol. 21, no. 2, pp. 195--212, 2024, doi: 10.1038/s41592-023-02151-z.

\bibitem{ref27}
A. Reinke et al., ``Understanding metric-related pitfalls in image analysis validation,'' \emph{Nat. Methods}, vol. 21, no. 2, pp. 182--194, 2024, doi: 10.1038/s41592-023-02150-0.

\bibitem{ref13}
D. S. Marcus, T. H. Wang, J. Parker, J. G. Csernansky, J. C. Morris, and R. L. Buckner, ``Open Access Series of Imaging Studies (OASIS): Cross-sectional MRI data in young, middle aged, nondemented, and demented older adults,'' \emph{J. Cogn. Neurosci.}, vol. 19, no. 9, pp. 1498--1507, 2007, doi: 10.1162/jocn.2007.19.9.1498.

\bibitem{ref14}
A. Hering et al., ``Learn2Reg: Comprehensive multi-task medical image registration challenge, dataset and evaluation in the era of deep learning,'' \emph{IEEE Trans. Med. Imaging}, vol. 42, no. 3, pp. 697--712, 2023, doi: 10.1109/TMI.2022.3213983.

\bibitem{ref15}
I. Loshchilov and F. Hutter, ``Decoupled weight decay regularization,'' in \emph{Proc. Int. Conf. Learn. Represent. (ICLR)}, 2019. [Online]. Available: \url{https://openreview.net/forum?id=Bkg6RiCqY7}

\bibitem{ref17}
L.-C. Chen, G. Papandreou, F. Schroff, and H. Adam, ``Rethinking atrous convolution for semantic image segmentation,'' arXiv:1706.05587, 2017. [Online]. Available: \url{https://arxiv.org/abs/1706.05587}

\bibitem{ref18}
M. M. Rahman and R. Marculescu, ``MK-UNet: Multi-kernel lightweight CNN for medical image segmentation,'' in \emph{Proc. IEEE/CVF Int. Conf. Comput. Vis. Workshops (ICCVW)}, 2025, pp. 1053--1062, doi: 10.1109/ICCVW69036.2025.00114.

\bibitem{ref19}
Z. Zhou, M. M. R. Siddiquee, N. Tajbakhsh, and J. Liang, ``UNet++: Redesigning skip connections to exploit multiscale features in image segmentation,'' \emph{IEEE Trans. Med. Imaging}, vol. 39, no. 6, pp. 1856--1867, 2020, doi: 10.1109/TMI.2019.2959609.

\bibitem{ref20}
E. Xie, W. Wang, Z. Yu, A. Anandkumar, J. M. Alvarez, and P. Luo, ``SegFormer: Simple and efficient design for semantic segmentation with transformers,'' in \emph{Adv. Neural Inf. Process. Syst.}, 2021, vol. 34, pp. 12077--12090. [Online]. Available: \url{https://proceedings.neurips.cc/paper/2021/hash/64f1f27bf1b4ec22924fd0acb550c235-Abstract.html}

\bibitem{ref21}
Y. Ou et al., ``Enhanced medical image segmentation via deep dynamic self-adjusting U-Net with multi-scale attention and semantic mitigation,'' \emph{Vis. Comput.}, vol. 41, no. 11, pp. 8385--8401, 2025, doi: 10.1007/s00371-025-03874-0.

\bibitem{ref22}
J. Chen et al., ``TransUNet: Rethinking the U-Net architecture design for medical image segmentation through the lens of transformers,'' \emph{Med. Image Anal.}, vol. 97, Art. no. 103280, 2024, doi: 10.1016/j.media.2024.103280.

\bibitem{ref23}
H. Cao et al., ``Swin-Unet: Unet-like pure transformer for medical image segmentation,'' in \emph{Computer Vision -- ECCV 2022 Workshops}, 2023, pp. 205--218, doi: 10.1007/978-3-031-25066-8\_9.

\end{thebibliography}
\end{document}